\documentclass[11pt]{article}
\usepackage{coling2018}
\usepackage{times}
\usepackage{url}
\usepackage{latexsym}

\usepackage{multirow}
\usepackage{graphicx}
\usepackage{subfig}
\usepackage{color}
\usepackage{titlesec}
\usepackage{wrapfig}
\usepackage{subfig}

\usepackage[group-separator={,}]{siunitx}

\usepackage{tikz,pgfplots}
\usepackage{array}

\titleformat{\paragraph}
{\normalfont\normalsize\bfseries}{\theparagraph}{1em}{}
\titlespacing*{\paragraph}
{0pt}{3.25ex plus 1ex minus .2ex}{1.5ex plus .2ex}

\title{Transfer Learning for British Sign Language Modelling}

\author{\kern-2em Boris Mocialov \\
  \kern-2em School of Mathematical \\ \kern-2em and Computer Sciences and \\
  \kern-2em Edinburgh Centre for Robotics \\
  \kern-2em Heriot-Watt University \\
  \kern-2em Edinburgh, UK \\
  {\kern-2em \tt bm4@hw.ac.uk} \\\And
  Graham Turner \\
  School of Social Sciences and \\
  Centre for Translation \& \\
  Interpreting Studies in Scotland \\
  Heriot-Watt University \\
  Edinburgh, UK \\
  {\tt g.h.turner@hw.ac.uk} \\\And
  \qquad Helen Hastie \\
  \qquad School of Mathematical \\ \qquad and Computer Sciences and \\
  \qquad Edinburgh Centre for Robotics \\
  \qquad Heriot-Watt University \\
  \qquad Edinburgh, UK \\
  {\qquad\tt h.hastie@hw.ac.uk} \\}
\date{}

\usepackage[pscoord]{eso-pic}

\newcommand{\placetextbox}[3]{
  \setbox0=\hbox{#3}
  \AddToShipoutPictureFG*{
    \put(\LenToUnit{#1\paperwidth},\LenToUnit{#2\paperheight}){\vtop{{\null}\makebox[0pt][c]{#3}}}%
  }%
}%

\begin{document}
\maketitle
\begin{abstract}


Automatic speech recognition and spoken dialogue systems have made great advances through the use of deep machine learning methods.  This is partly due to greater computing power but also through the large amount of data available in common languages, such as English. Conversely, research in minority languages, including sign languages, is hampered by the severe lack of data. This has led to work on transfer learning methods, whereby a model developed for one language is reused as the starting point for a model on a second language, which is less resourced. In this paper, we examine two transfer learning techniques of fine-tuning and layer substitution for language modelling of British Sign Language. Our results show improvement in perplexity when using transfer learning with standard stacked LSTM models, trained initially using a large corpus for standard English from the Penn Treebank corpus.


\end{abstract}
\blfootnote{
    %
    %
    %
    %
     \hspace{-0.65cm}  
     This work is licenced under a Creative Commons 
     Attribution 4.0 International Licence.
     Licence details:
     \url{http://creativecommons.org/licenses/by/4.0/}
    %
    %
}

\section{Introduction}

Spoken dialogue systems and voice assistants have been developed to facilitate natural conversation between machines and humans. They provide services through devices such as Amazon Echo Show and smartphones to help the user do tasks \cite{mctear:2004} and, more recently, for open domain chitchat \cite{serban2016generative}, all through voice.
Recent advances have been facilitated by the huge amounts of data collected through such devices and have resulted in the recent success of deep machine methods, providing significant improvements in performance. However, not all languages are able to benefit from these advances, particularly those that are under-resourced. These include sign languages and it means that those who sign are not able to leverage such interactive systems nor the benefits that automatic transcription and translating of signing would afford. 

Here, we advance the state of the art with respect to transcribing British Sign Language (BSL). Our aim is for automated transcription of the BSL into English leveraging video recognition technologies. BSL enables communication of meaning through parameters such as hand shape, position, hand orientation, motion, and non-manual signals \cite{sutton1999linguistics}. BSL has no standard notation for writing the signs, as with letters and words in English.
Analogous to the International Phonetic Alphabet (IPA), highly detailed mapping of visual indicators to written form are available, such as HamNoSys \cite{hanke2004hamnosys}. Despite the expressiveness of the HamNoSys writing system, its practical uses are limited and only a handful of experts know how to use it. Recent methods for automatic speech recognition (ASR) use deep neural models to  bypass the need for phoneme dictionaries \cite{DBLP:journals/corr/HannunCCCDEPSSCN14}, which are then combined with language models. 
\sloppy Previous work \cite{mocialov2016towards,mocialovtowards} has shown that we can use visual features to automatically predict individual signs. This work follows on in that these individual signs are to be used with a language model to take into account context and therefore increase accuracy of the transcriber, which outputs a string of word-like tokens. These tokens are called glosses \cite{sutton1999linguistics,cormier2015bsl}. Although glosses are translated BSL signs, they also convey some grammatical information about BSL. This makes glosses useful in their own right without the videos of the BSL signs and sheds some light into the syntax and semantics of the BSL. 





This paper focuses on language modelling, a common technique in the field of ASR and Natural Language Processing to model the likelihood of certain words following each other in a sequence. We improve modelling of the BSL glosses by proposing to use transfer learning approaches, such as fine-tuning and layer substitution. 
The use of transfer learning technique can overcome the data sparsity issue in statistical modelling for scarce resource languages by using similar resources that can be found in large quantities and then further training the models on a specific low resource data.

We show that a model, pre-trained on the  Penn Treebank (PTB) dataset\footnote{https://catalog.ldc.upenn.edu/ldc99t42} and fine-tuned on the BSL monolingual corpus\footnote{http://www.bslcorpusproject.org/} can yield better results. This is in contrast to the same architecture that is trained directly on the BSL dataset without pre-training.
This is a somewhat surprising result as there are marked differences between the two languages, particularly with the respect to the syntax \cite{sutton1999linguistics}. 

The paper begins with presenting methods for modelling languages and how they can be utilised in the BSL modelling. Section~\ref{transfer_learning} gives an overview of how transfer learning can be achieved as well as the use of transfer learning in sign languages. Section~\ref{datadatadata} gives an overview of the datasets that are used in this paper, their statistics, and pre-processing steps to create two monolingual corpora for statistical model training. Section~\ref{methodologymethodology} describes in detail the setup for the experiments in this paper. Section~\ref{resultsresults} presents the results of the models employed for this research and discusses these results and the limitations of the approach taken in terms of the data used in Section~\ref{discussion}. The paper is then concluded and future work is proposed.

\section{Related Work}

\subsection{Sign Language Modelling}\label{sign_language}

Despite the availability of many alternatives for language modelling, such as count-based n-grams and their variations \cite{chen1999empirical,rosenfeld2000two,maccartney2005nlp,bulyko2007language,guthrie2006closer}, hidden Markov models \cite{dreuw2008visual,dreuw2008benchmark}, decision trees and decision forests \cite{filimonov2011decision}, and neural networks \cite{deena2016combining,mikolov2010recurrent}, research in sign language modelling predominantly employs simple n-gram models, such as in \newcite{DBLP:journals/corr/CateH17}, \newcite{forster2012rwth}, and \newcite{masso2010dealing}. 

The reason for the wide-spread use of n-grams in sign language modelling is the simplicity of the method. However, there is a disconnect between n-grams and sign language in that signing is embodied and perceived visually, while the n-grams are commonly applied to text sequence modelling. For this reason, the authors in \newcite{stein2007hand}, \newcite{zhao2000machine}, \newcite{dreuw2008benchmark}, \newcite{masso2010dealing}, and \newcite{forster2013improving} model glosses, such as the ones shown on Figure~\ref{elanannot}, which are obtained from the transcribed sign languages, in a similar way to how language modelling is applied to automatic transcribed words from speech. 


Glosses model the meaning of a sign in a written language, but not the execution (i.e. facial expressions, hand movement). Therefore, the more detailed meaning of what was signed may get lost when working with the higher-level glosses. To overcome this issue and to incorporate valuable information into sign language modelling, additional features are added in similar research, such as non-manual features (e.g facial expressions) \cite{san2009spoken,masso2010dealing,zhao2000machine,stein2007hand}.

In this work we use glosses because we want to model BSL purely at the gloss level without any additional information (e.g. facial expressions).

\subsection{Transfer Learning}\label{transfer_learning}

While transfer learning is a more general machine learning term, cross-domain adaptation of language models is used in the language modelling literature \cite{deena2016combining,ma2017approaches}. Models are usually trained on some specific domain that consists of a specific topic, genre, and similar features that can be identified by an expert. For example, a restaurant domain when a new type of a restaurant is created then the system needs to be able to adapt and be able to understand and discuss this new type of the restaurant. Unfortunately, it is nearly impossible to train a model for all possible configuration of current or future features. 
%
Commonly, a set of features are extracted from the raw data. When features change, re-training is required. 
Useful features can also be extracted without expert knowledge with such techniques as Latent Dirichlet Allocation (LDA). These features usually take the form of words that represent topics in the data \cite{deena2016combining}. Best practice tries to avoid re-training the models every time one of the features changes as the domain changes due to the overhead involved.

Model-based adaptation to the new domains, on the other hand, is achieved by either fine-tuning or the introduction of adaptation layer(s) \cite{yosinski2014transferable}. Fine-tuning involves further training the already pre-trained model using the data from the new domain. The intuition behind the fine-tuning is that it is much quicker to learn new information with related knowledge. The adaptation layer approach incorporates new knowledge by re-training only the adaptation layer, whereas the rest of the model remains exactly the same as if it was used in the original domain and acts as a feature extractor for the new domain \cite{deena2016combining}.

Transfer learning has been applied to sign languages in computing for various purposes to demonstrate that the method is suitable for the task due to the lack of substantial domain-specific sign language data. Transfer learning has been successfully applied to static pose estimation, transferring the knowledge from pose estimation to the sign language pose estimation \cite{DBLP:journals/corr/GattupalliGA16} and classification of fingerspelled letters in American Sign Language \cite{garcia2016real,karthickaryajayeshkudase2017,belalchaudhary2017,muskandhimandrg.n.rathna2017}. In particular, most of the transfer learning in sign language has been applied to static image recognition to recognise the hand shape in an image using convolutional neural networks.

We apply transfer learning to the language modelling task as this is a key challenge in successfully transcribing BSL. 
%

\section{Corpora}\label{datadatadata}

The BSL corpus and the preprocessed Penn Treebank (PTB) corpus were chosen for this research. The monolingual PTB dataset consists of telephone speech, newswire, microphone speech, and transcribed speech. The dataset is preprocessed to eliminate letters, numbers, or punctuation and was used by Mikolov~\shortcite{mikolov2010recurrent}. The BSL corpus contains video conversations among deaf native, near-native and fluent signers across the United Kingdom. Almost all of the approximately one hundred recorded conversations are annotated for thirty seconds each at the gloss level using ELAN\footnote{https://tla.mpi.nl/tools/tla-tools/elan/} annotation tool \cite{schembri2013building}.

\begin{figure}[h!]
\centering
\includegraphics[width=0.4\linewidth]{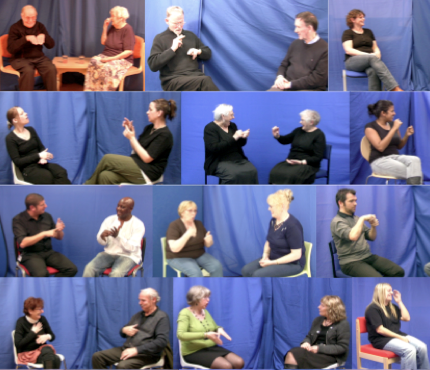}
\caption{The BSL Corpus Project Sample Video Snippets\protect\footnotemark}
\label{bslcorpus}
\end{figure}
\footnotetext{http://www.bslcorpusproject.org/cava/}

All recordings of the signers were made using up to four standard video cameras with a plain backdrop to provide full body view of the individuals, as well as, views from above of their use of signing
space. The conversations between the signers included signing personal experience anecdotes, spontaneous conversations \cite{schembri2013building}.

The BSL data that we focused on was narratives between two participants, where one person had to think of a topic to sign about to another participant during the elicitation.

\placetextbox{0.1}{0.77}{\textsf{a)}}%
\placetextbox{0.1}{0.705}{\textsf{b)}}%
\placetextbox{0.1}{0.675}{\textsf{c)}}%

\begin{figure}[h!]
\centering
\includegraphics[width=\linewidth,trim={0 0 20cm 17cm},clip]{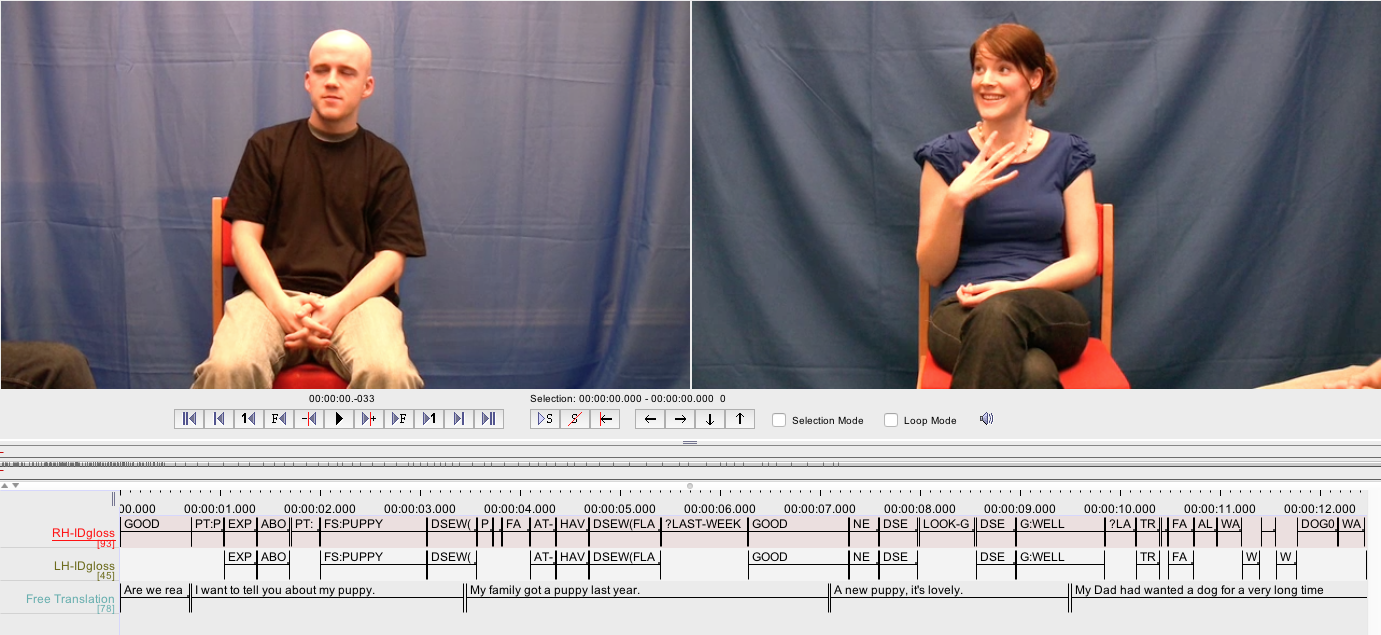}
\end{figure}
\vspace{-1.5em}
\begin{table}[h!]
\centering
\label{my-label}
\resizebox{\textwidth}{!}{%
\begin{tabular}{|l|l|l|l|l|l|l|l|l|l|l|l|l|l|l|}
\hline
\textbf{RH-IDgloss}       & PT:PRO1SG                         & EXPLAIN & ABOUT & PT:POSS1SG & FS:PUPPY & DSEW(FLAT)-BE:ANIMAL & PT:POSS1SG   & WANT   & FAMILY   & AT-LAST   & HAVE  & DSEW(FLAT)-BE:ANIMAL & ?LAST-WEEK & GOOD  \\ \hline
\textbf{LH-IDgloss}       &                                   & EXPLAIN & ABOUT &            & FS:PUPPY & DSEW(FLAT)-BE:ANIMAL &              &        &          & AT-LAST          & HAVE      & DSEW(FLAT)-BE:ANIMAL & & GOOD \\ \hline
\textbf{Free Translation} & \multicolumn{6}{c|}{I want to tell you about my puppy}                      & \multicolumn{8}{c|}{My family got a puppy last year}  \\ \hline
\multicolumn{15}{c}{}   \\
\multicolumn{15}{c}{}   \\
\hline
\textbf{Model Input Gloss}       &                          & EXPLAIN & ABOUT &  & PUPPY & ANIMAL &    &  WANT  &  FAMILY  &  AT-LAST  &  HAVE & ANIMAL & LAST-WEEK &  GOOD \\ \hline
\end{tabular}
}
\end{table}

\begin{figure}[h!]
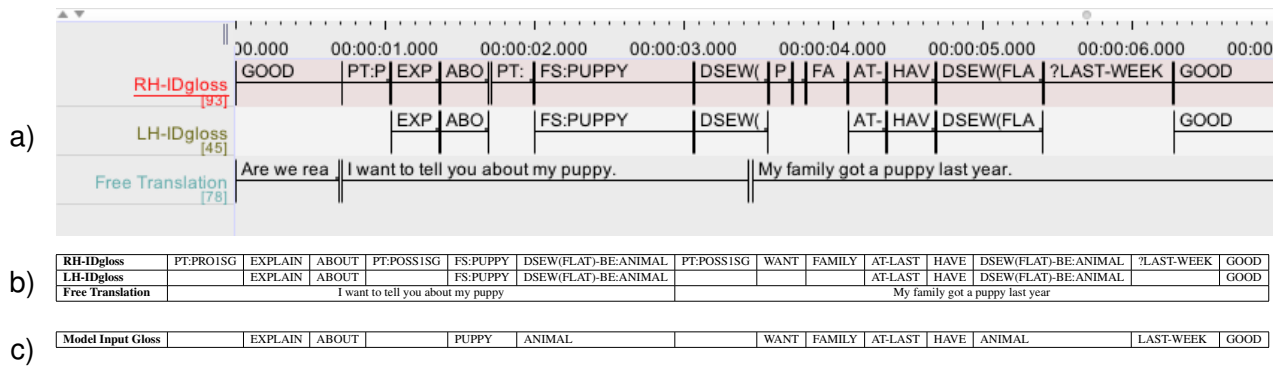

\centering
\caption{a) The BSL Corpus Annotation in ELAN; b) Table shows full text of the annotated glosses for the two first sentences from the ELAN annotation; c) Glosses that are used for the BSL modelling}
\label{elanannot}
\end{figure}

The corpus is annotated with glosses, taken from the BSL SignBank in ELAN as shown in Figure~\ref{elanannot}a. Figure~\ref{elanannot}b shows all the glosses of the first sentence. As mentioned above, gloss is an identifier of a unique sign, written in English and should represent its phonological and morphological meaning \cite{schembri2013building}. In the corpus, the glosses are identified throughout the videos for both left and right hands as sometimes different signs can be signed at the same time. Apart from the glossing, the annotations include the corresponding free English written translation of the meaning of the signing split into sentences (see the Free Translation in the Figure~\ref{elanannot}). Figure~\ref{elanannot}c shows which glosses are considered for the BSL modelling and which are ignored. This is done to match the vocabulary of the PTB corpus for the transfer learning purposes.

\subsection{Data Pre-processing}\label{datapreproc}

For the BSL corpus, we ignore the free translation and extract English text from the glosses, preserving the order of the signs executed. For example, in Figure~\ref{elanannot}, right-hand glosses identify the following order of the signs: good, explain, about, puppy, etc. excluding glosses, such as PT:PRO for pointing signs or PT:POSS for possessives and others (Figure~\ref{elanannot}c), which are explained in more detail in Fenlon et al.~\shortcite{fenlon2014using}. Since the gloss annotation does not include explicit punctuation, it is impossible to tell where a signed sentence begins and where it stops. To overcome this limitation of the gloss annotation, we use the Free Translation annotation, which gives the boundaries of sentences in videos. Later, we split the extracted glosses into sentences using these sentence boundaries. By the end of the pre-processing stage, we have glosses (excluding special glosses for pointing signs, posessives or other non-lexical glosses) in the order that the corresponding signs were executed in the video, split into sentences. As a result, we extracted 810 nominal sentences from the BSL corpus with an average length of the sentence being 4.31 glossed signs, minimum and maximum lengths of 1 and 13 glossed signs respectively. A monolingual dataset has been created with the extracted sentences. As obtained from the PTB dataset \cite{merityRegOpt}, the English language corpus has 23.09 words on average per sentence with minimum being 3 and maximum 84 words per sentence. The pre-processed BSL corpus has a vocabulary of 666 words, while the PTB dataset has a vocabulary of \num[group-separator={,}]{10002} words. From this point on in this paper, we will use the term `words' to refer to both glosses in the BSL and words in the PTB datasets because we aim to use a common vocabulary for training our models.

Both monolingual datasets were split into training, validation, and testing sets as required for training and evaluation of the statistical models. Both datasets were split using ratio 85:15. The smaller subset, in turn, was split 50:50 for validation and testing for the two datasets.

\section{Language Modelling Methodology}\label{methodologymethodology}

\subsection{Statistical Language Models}

Perplexity measure has been used for evaluation and comparison purposes of different models. We used the following formula to calculate the perplexity values: $e^{Cross-Entropy}$ as used in \newcite{bengio2003neural}, which approximates geometric average of the predicted words probabilities on the test set. We have explicitly modelled out-of-vocabulary (OOV), such as $<unk>$ placeholder in all the experiments.

\subsubsection{Neural Models}

For comparison, we use two methods: 1) stacked LSTM and 2) Feed-Forward (FFNN) architectures to create the BSL language models. All models are implemented in PyTorch\footnote{http://pytorch.org/} with weight-drop recurrent regularisation scheme for the LSTMs, which is important for overcoming commonly known LSTM model generalisation issues \cite{merityRegOpt,merityAnalysis}. The feed-forward model, on the other hand, had no regularisations as it is less susceptible to overfitting due to the much smaller number of parameters.

The parameters that were modified to achieve the lowest perplexity were input size of the overall input sequence for the recurrent neural network (back-propagation through time, BPTT), batch size, learning rate, and the optimizer. The parameters were selected using the grid search approach using perplexity metric. As a result, for the stacked LSTMs, bptt was set to 5, batch size was set to 16, discounted learning rate was set to 30, and the optimizer was set to stochastic gradient descent. In case of the feed-forward network, input was set to 5 words, batch size was set to 16, discounted learning rate was set to 30, and the optimizer was set to stochastic gradient descent. All the neural models were trained for 100 epochs.

In the case of the neural networks, the sequences of words were tokenised (i.e. turned into integers) and the tokenisation was stored to ensure the same tokenisation during the transfer learning phase. The input, therefore, consisted of a set of tokens, while the outputs (i.e. predicted words) were turned into a one-hot vectors.

\begin{figure}[h]
    \centering
    \subfloat[\mbox{Stacked LSTMs model}]{\label{stackedlstmsgraph}{\scalebox{0.4}{\includegraphics[width=0.4\linewidth]{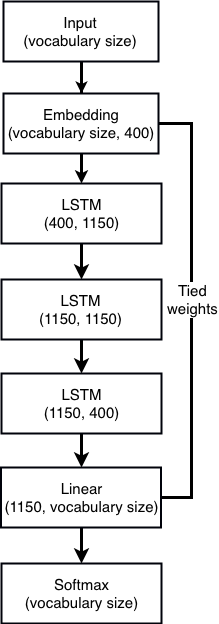} }}}%
    \qquad\qquad\qquad
    \subfloat[\mbox{Feed-Forward model}]{\label{ffnngraph}{\scalebox{0.3}{\includegraphics[width=0.4\linewidth]{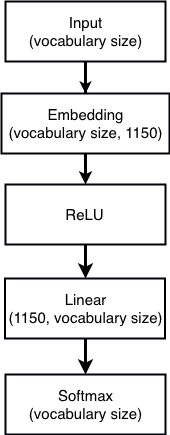} }}}%
    \caption{The two types of neural models used to test transfer methods for sign language modelling}%
    \label{fig:example}%
\end{figure}

\paragraph{Stacked LSTMs}
Figure~\ref{stackedlstmsgraph} shows the architecture of the stacked LSTM model. The model consists of an embedding layer of 400 nodes, which, together with the tokenisation, turns string of words into a vector of real numbers. Secondly, three LSTM layers with 1150 nodes each are stacked vertically for deeper feature extraction. Thirdly, the linear layer downsizes the stacked LSTMs output to the vocabulary size and applies linear transformation with softmax normalisation. The weights of the embedding and the linear layers are tied. This means that the two layers share the same weights, which reduces the number of parameters of the network and makes the convergence during training faster. The same architecture was used in \newcite{merityRegOpt} to model PTB dataset, reporting 57.3 perplexity, utilising cache in the model from recent predictions.

\paragraph{FFNN}
Figure~\ref{ffnngraph} shows the Feed-forward model architecture. The model does not have the stacked LSTMs layers. Instead, the stacked LSTMs are substituted with one hidden fully-connected rectifier layer, which is known to overcome the vanishing gradient problem. The weights of the embedding and the outputs layers are not tied together. Similar architectures have been used for language modelling in \newcite{le2013structured}, \newcite{4960686}, and \newcite{DBLP:journals/corr/BrebissonSAVB15} with the hidden layer having different activation functions with the PTB dataset being used in \newcite{DBLP:journals/corr/AudhkhasiSR14}, reporting 137.32 perplexity.

%

\subsubsection{Training the Models}

Transfer learning was achieved with both fine-tuning and substitution. Both FFNN and LSTM were trained on the PTB dataset and then either fine-tuned or the last layer was substituted with the new adaptation layer, freezing the rest of the weights, and further training on the BSL dataset.

To achieve fine-tuning, first the best model is saved after the training of both the FFNN and the stacked LSTMs on the PTB dataset. Then the training is restarted on the BSL corpus, having initialised the model with the weights, trained on the PTB dataset. 

To perform layer substitution as a transfer learning approach, the same first step as with the fine-tuning is repeated and the model, trained on the PTB, is saved. When the training is restarted on the BSL dataset, the saved model is loaded and the last linear layer is substituted with a layer that has as many nodes as the BSL vocabulary. Later, all the weights of the network are locked and will not be modified during the optimisation. Only the weights of the last substituted layer will be modified. This method uses the pretrained network as a feature extractor and only modifies the last layer weights to train the model for the BSL dataset.


\section{Results}\label{resultsresults}
This section is split into two subsections. We firstly present results without transfer learning, namely both the FFNN and the stacked LSTMs models trained and tested on the PTB dataset or trained and tested on the BSL. Later we present results with the transfer learning, with both FFNN and the stacked LSTMs models trained on the PTB dataset and then fine-tuned and tested on the BSL.

To show that the two languages are different, as discussed in Section~\ref{datapreproc}, we applied the model trained on one language to the other language and vice versa. As a result, the model trained on English language and applied to the BSL scored 1051.91 in perplexity using SRILM toolkit \cite{stolcke2002srilm}. Conversely, a model trained on the BSL has been applied to the English language and scored 1447.23 in perplexity. As expected, the perplexity is high in both cases, which means that the probability distribution over the next word in one language is far from the true distribution of words in the other language. 


\subsection{Without Transfer Learning}

\begin{table}[!h]
\centering
\resizebox{0.5\textwidth}{!}{%
\setlength{\tabcolsep}{0.5em} 
{\renewcommand{\arraystretch}{2.0}
\begin{tabular}{|c|c|c|c|}
\cline{1-4}
\textbf{Method}                                                                        & \textbf{\begin{tabular}[c]{@{}c@{}}Penn Treebank\\(PTB)\end{tabular}} & \multicolumn{2}{c|}{\textbf{\begin{tabular}[c]{@{}c@{}}The BSL\\Corpus Project\end{tabular}}} \\ \cline{1-4}

\begin{tabular}[c]{@{}c@{}}FFNN\end{tabular}                               & 190.46 & \multicolumn{2}{c|}{\textbf{258.1}} \\ \cline{1-4}
\begin{tabular}[c]{@{}c@{}}Stacked LSTMs\end{tabular}                               & 65.91 & \multicolumn{2}{c|}{274.03} \\ \cline{1-4}

OOV              & 6.09\% & \multicolumn{2}{c|}{25.18\%} \\ \hline

\end{tabular}%
}
}
\caption{\label{perplexities_table}Perplexities on either the PTB or the BSL test sets using models trained and tested on the same corpus (i.e. PTB and BSL)}
\end{table}

Table~\ref{perplexities_table} shows perplexities on the two datasets with two statistical models. From the table, we can infer that the trained models on the PTB dataset have lower perplexity than the same architectures trained on the BSL dataset. This can be explained by the fact that the PTB dataset has more data than the BSL dataset and, therefore, statistical models can generalise better. Furthermore, the amount of data is further reduced in the BSL case as the OOV covers a quarter of the overall dataset.


\subsection{With Transfer Learning}

Table~\ref{perplexities_table2} shows perplexities on the two datasets with two statistical models, applying transfer learning. From this table, it can be seen that the substitution approach gives very similar results independent of the whether FFNN or stacked LSTMs model is used (123.92 versus 125.32). The best result is achieved with the fine-tuning approach on the stacked LSTMs model, while the higher perplexity result is on the FFNN model with the fine-tuning approach. Similar results have been reported in \newcite{irie2016lstm}, where fine-tuned GRU performed worse than fine-tuned LSTM model. In addition, the OOV count differs from that of the Table~\ref{perplexities_table} due to the fact that a subset of the vocabulary, observed in the PTB dataset during training is then identified in the BSL dataset during testing.

\begin{table}[!h]
\centering
\resizebox{0.5\textwidth}{!}{%
\setlength{\tabcolsep}{0.5em} 
{\renewcommand{\arraystretch}{1.2}
\begin{tabular}{|c|c|c|c|c|}
\hline

\tiny \textbf{\begin{tabular}[c]{@{}c@{}} \qquad \\ Method\\ \qquad \end{tabular}} & \multicolumn{2}{c|}{\textbf{\tiny Fine-tuning}}  & \multicolumn{2}{c|}{\textbf{\tiny Substitution}} \\ \cline{1-1}  \hline

\tiny \begin{tabular}[c]{@{}c@{}}FFNN\end{tabular} &  \multicolumn{2}{c|}{\tiny 179.3} & \multicolumn{2}{c|}{\tiny 123.92} \\ \cline{1-1} \hline

\tiny \begin{tabular}[c]{@{}c@{}}Stacked LSTMs\end{tabular}                    & \multicolumn{2}{c|}{\tiny \textbf{121.46}} & \multicolumn{2}{c|}{\tiny 125.32} \\ \hline

\multicolumn{1}{|c|}{\tiny OOV}              & \multicolumn{4}{c|}{\tiny 12.71\%}  \\ \hline

\end{tabular}%
}
}
\caption{\label{perplexities_table2}Perplexities on the BSL test set after applying the transfer learning on FFNN and LSTMs}

\end{table}

\subsection{Discussion}\label{discussion}

The salient idea of this paper is whether transfer learning is a legitimate method for modelling one language with the knowledge of another, assuming the languages are different, but share some common properties, such as vocabulary. This theory is intuitive and has been discussed in linguistics for spoken languages \cite{kaivapalu2007morphology}. In our case, PTB corpus covers most of the vocabulary found in the BSL corpus (12.71\% OOV) by the virtue of the gloss annotation of the BSL corpus \cite{schembri2013building}. However, the languages are assumed to be different as they evolved independently of one another \cite{faberfaber}.

The results obtained are different from reported in similar research. For example, for the FFNN model, Audhkhasi et al.~\shortcite{DBLP:journals/corr/AudhkhasiSR14} report 137.32 versus our achieved 190.46 perplexity and for the stacked LSTMs model, Merity et al.~\shortcite{merityRegOpt} report 57.3 versus our achieved 65.91 perplexity. This can be explained by the fact that not all the regularisation techniques had been used in this research as in the past research and the model training had been restricted to 100 epochs. Further training may further reduce the perplexity to that reported in Merity et al.~\shortcite{merityRegOpt}.

From the results, we can see that the transfer learning leads to superior models than the models trained on the BSL directly (258.1 and 274.03 against 123.92 and 125.32). Since the quality of the trained models using either of the approaches is similar in case of the stacked LSTMs model (121.46 and 125.32), the choice between the fine-tuning and substitution can be guided based on the convergence speed. During the substitution, only one layer of the network is replaced with a new one and the rest of the weights in the network are locked, therefore, one set of weights will be optimized. This is in contrast to the fine-tuning method, which optimizes all of the weights, which may, in turn, require more interactions, depending on how different the new data is.
%

\section{Conclusion}

This paper shows how transfer learning techniques can be used to improve language modelling for the BSL language at the gloss level. Statistical modelling techniques are used to generate language models and to evaluate them using a perplexity measure.

The choice of the transfer learning technique is guided by the scarcity of available resources of the BSL language and the availability of the English language dataset that shares similar language modelling vocabulary with the annotated BSL. Feed-forward and recurrent neural models have been used to evaluate and compare generated language models. The results show that transfer learning can achieve superior quality of the generated language models. However, our pre-processed BSL corpus lacks constructs that are essential for a sign language, such as classifier signs and others. Nevertheless, transfer learning for modelling the BSL shows promising results and should be investigated further.

\subsection{Future Work}

Although this paper discusses the use of a model initially trained on English and presents promising preliminary results, the annotation of the BSL, used in this paper, is limited as this paper serves as a proof of concept. In particular, the annotation used is missing some of the grammatical aspects of the BSL, such as classifier signs and others. Inclusion of these into the BSL language modelling would increase the OOV count as the English language does not have equivalent language constructs. This raises a question whether a sign language can be modelled using other languages that may have these constructs. More generally, is it possible to model a language with transfer learning using other less-related languages? Similar questions have been partly answered for the written languages in the field of machine translation \cite{gu2018universal} by bringing words of different languages close to each other in the latent space. However, nothing similar has been done for the sign languages. 

From the methodological side of the modelling, additional advanced state of the art techniques should be experimented with to achieve greater quality of the generated models, such as attention mechanism for the recurrent neural networks. Finally, this paper focuses on key techniques for sign processing, which could be part of a larger conversational system whereby signers could interact with computers and home devices through their natural communication medium of sign. 
Research in such end-to-end systems would include vision processing, segmentation, classification, and language modelling as well as language understanding and dialogue modelling, all tuned to sign language. 

%
%
%
    %
    %
    %
    %
    %
    %


\bibliographystyle{acl}
\bibliography{bibl,sds,conll2018}

\begin{thebibliography}{}

\bibitem[\protect\citename{Audhkhasi \bgroup et al.\egroup
  }2014]{DBLP:journals/corr/AudhkhasiSR14}
Kartik Audhkhasi, Abhinav Sethy, and Bhuvana Ramabhadran.
\newblock 2014.
\newblock Diverse embedding neural network language models.
\newblock {\em arXiv preprint arXiv:1412.7063}, abs/1412.7063.

\bibitem[\protect\citename{Bengio \bgroup et al.\egroup
  }2003]{bengio2003neural}
Yoshua Bengio, R{\'e}jean Ducharme, Pascal Vincent, and Christian Jauvin.
\newblock 2003.
\newblock A neural probabilistic language model.
\newblock {\em Journal of machine learning research}, 3(Feb):1137--1155.

\bibitem[\protect\citename{Brennan}1992]{faberfaber}
Mary Brennan.
\newblock 1992.
\newblock {\em The visual world of BSL: An introduction}.
\newblock Faber and Faber.
\newblock In David Brien (Ed.), Dictionary of British Sign Language/English.

\bibitem[\protect\citename{Bulyko \bgroup et al.\egroup
  }2007]{bulyko2007language}
Ivan Bulyko, Spyros Matsoukas, Richard Schwartz, Long Nguyen, and John Makhoul.
\newblock 2007.
\newblock Language model adaptation in machine translation from speech.
\newblock In {\em Proceedings of the Acoustics, Speech and Signal Processing,
  2007. ICASSP 2007. IEEE International Conference on}, volume~4, pages
  IV--117. IEEE.

\bibitem[\protect\citename{Cate and Hussain}2017]{DBLP:journals/corr/CateH17}
Hardie Cate and Zeshan Hussain.
\newblock 2017.
\newblock Bidirectional american sign language to english translation.
\newblock {\em arXiv preprint arXiv:1701.02795}, abs/1701.02795.

\bibitem[\protect\citename{Chaudhary}2017]{belalchaudhary2017}
Belal Chaudhary.
\newblock 2017.
\newblock Real-time translation of sign language into text.
\newblock Data Science Retreat. https://github.com/BelalC/sign2text, apr.

\bibitem[\protect\citename{Chen and Goodman}1999]{chen1999empirical}
Stanley~F Chen and Joshua Goodman.
\newblock 1999.
\newblock An empirical study of smoothing techniques for language modeling.
\newblock {\em Computer Speech \& Language}, 13(4):359--394.

\bibitem[\protect\citename{Cormier \bgroup et al.\egroup }2015]{cormier2015bsl}
Kearsy Cormier, Jordan Fenlon, Sannah Gulamani, and Sandra Smith.
\newblock 2015.
\newblock Bsl corpus annotation conventions.

\bibitem[\protect\citename{de Br{\'{e}}bisson \bgroup et al.\egroup
  }2015]{DBLP:journals/corr/BrebissonSAVB15}
Alexandre de~Br{\'{e}}bisson, {\'{E}}tienne Simon, Alex Auvolat, Pascal
  Vincent, and Yoshua Bengio.
\newblock 2015.
\newblock Artificial neural networks applied to taxi destination prediction.
\newblock {\em arXiv preprint arXiv:1508.00021}, abs/1508.00021.

\bibitem[\protect\citename{Deena \bgroup et al.\egroup
  }2016]{deena2016combining}
Salil Deena, Madina Hasan, Mortaza Doulaty, Oscar Saz, and Thomas Hain.
\newblock 2016.
\newblock Combining feature and model-based adaptation of rnnlms for
  multi-genre broadcast speech recognition.
\newblock In {\em Proceedings of the Annual Conference of the International
  Speech Communication Association, INTERSPEECH}, pages 2343--2347. Sheffield.

\bibitem[\protect\citename{Dreuw and Ney}2008]{dreuw2008visual}
Philippe Dreuw and Hermann Ney.
\newblock 2008.
\newblock Visual modeling and feature adaptation in sign language recognition.
\newblock In {\em Voice Communication (SprachKommunikation), 2008 ITG
  Conference on}, pages 1--4. VDE.

\bibitem[\protect\citename{Dreuw \bgroup et al.\egroup
  }2008]{dreuw2008benchmark}
Philippe Dreuw, Carol Neidle, Vassilis Athitsos, Stan Sclaroff, and Hermann
  Ney.
\newblock 2008.
\newblock Benchmark databases for video-based automatic sign language
  recognition.
\newblock In {\em LREC}.

\bibitem[\protect\citename{Fenlon \bgroup et al.\egroup }2014]{fenlon2014using}
Jordan Fenlon, Adam Schembri, Ramas Rentelis, David Vinson, and Kearsy Cormier.
\newblock 2014.
\newblock Using conversational data to determine lexical frequency in {British
  Sign Language}: The influence of text type.
\newblock {\em Lingua}, 143:187--202.

\bibitem[\protect\citename{Filimonov}2011]{filimonov2011decision}
Denis Filimonov.
\newblock 2011.
\newblock {\em Decision tree-based syntactic language modeling}.
\newblock University of Maryland, College Park.

\bibitem[\protect\citename{Forster \bgroup et al.\egroup
  }2012]{forster2012rwth}
Jens Forster, Christoph Schmidt, Thomas Hoyoux, Oscar Koller, Uwe Zelle,
  Justus~H Piater, and Hermann Ney.
\newblock 2012.
\newblock Rwth-phoenix-weather: A large vocabulary sign language recognition
  and translation corpus.
\newblock In {\em Proceedings of the 8th International Conference on Language
  Resources and Evaluation, LREC}.

\bibitem[\protect\citename{Forster \bgroup et al.\egroup
  }2013]{forster2013improving}
Jens Forster, Oscar Koller, Christian Oberd{\"o}rfer, Yannick Gweth, and
  Hermann Ney.
\newblock 2013.
\newblock Improving continuous sign language recognition: Speech recognition
  techniques and system design.
\newblock In {\em Proceedings of the Fourth Workshop on Speech and Language
  Processing for Assistive Technologies}, pages 41--46.

\bibitem[\protect\citename{Garcia and Viesca}2016]{garcia2016real}
Brandon Garcia and Sigberto~Alarcon Viesca.
\newblock 2016.
\newblock Real-time american sign language recognition with convolutional
  neural networks.
\newblock In {\em In Proceedings of Machine Learning Research, pp. 225-232.}

\bibitem[\protect\citename{Gattupalli \bgroup et al.\egroup
  }2016]{DBLP:journals/corr/GattupalliGA16}
Srujana Gattupalli, Amir Ghaderi, and Vassilis Athitsos.
\newblock 2016.
\newblock Evaluation of deep learning based pose estimation for sign language.
\newblock {\em arXiv preprint arXiv:1602.09065}, abs/1602.09065.

\bibitem[\protect\citename{Gu \bgroup et al.\egroup }2018]{gu2018universal}
Jiatao Gu, Hany Hassan, Jacob Devlin, and Victor~OK Li.
\newblock 2018.
\newblock Universal neural machine translation for extremely low resource
  languages.
\newblock {\em arXiv preprint arXiv:1802.05368}.

\bibitem[\protect\citename{Guthrie \bgroup et al.\egroup
  }2006]{guthrie2006closer}
David Guthrie, Ben Allison, Wei Liu, Louise Guthrie, and Yorick Wilks.
\newblock 2006.
\newblock A closer look at skip-gram modelling.
\newblock In {\em Proceedings of the 5th international Conference on Language
  Resources and Evaluation (LREC-2006)}.

\bibitem[\protect\citename{Hanke}2004]{hanke2004hamnosys}
Thomas Hanke.
\newblock 2004.
\newblock Hamnosys-representing sign language data in language resources and
  language processing contexts.

\bibitem[\protect\citename{Hannun \bgroup et al.\egroup
  }2014]{DBLP:journals/corr/HannunCCCDEPSSCN14}
Awni~Y. Hannun, Carl Case, Jared Casper, Bryan Catanzaro, Greg Diamos, Erich
  Elsen, Ryan Prenger, Sanjeev Satheesh, Shubho Sengupta, Adam Coates, and
  Andrew~Y. Ng.
\newblock 2014.
\newblock Deep speech: Scaling up end-to-end speech recognition.
\newblock {\em CoRR}, abs/1412.5567.

\bibitem[\protect\citename{Irie \bgroup et al.\egroup }2016]{irie2016lstm}
Kazuki Irie, Zoltan Tuske, Tamer Alkhouli, Ralf Schluter, and Hermann Ney.
\newblock 2016.
\newblock {LSTM}, gru, highway and a bit of attention: an empirical overview
  for language modeling in speech recognition.
\newblock Technical report, RWTH Aachen University Aachen Germany.

\bibitem[\protect\citename{Kaivapalu and Martin}2007]{kaivapalu2007morphology}
Annekatrin Kaivapalu and Maisa Martin.
\newblock 2007.
\newblock {Morphology} in {Transition}: {Plural} {Inflection} of {Finnish}
  nouns by {Estonian} and {Russian} {Learners}.
\newblock {\em Acta Linguistica Hungarica}, 54(2):129--156.

\bibitem[\protect\citename{Karthick~Arya}2017]{karthickaryajayeshkudase2017}
Jayesh~Kudase Karthick~Arya.
\newblock 2017.
\newblock Convolutional neural networks based sign language recognition.
\newblock {\em International Journal of Innovative Research in Computer and
  Communication Engineering}, 5(10), oct.

\bibitem[\protect\citename{Le \bgroup et al.\egroup }2013]{le2013structured}
Hai-Son Le, Ilya Oparin, Alexandre Allauzen, Jean-Luc Gauvain, and
  Fran{\c{c}}ois Yvon.
\newblock 2013.
\newblock Structured output layer neural network language models for speech
  recognition.
\newblock {\em IEEE Transactions on Audio, Speech, and Language Processing},
  21(1):197--206.

\bibitem[\protect\citename{Ma \bgroup et al.\egroup }2017]{ma2017approaches}
Min Ma, Michael Nirschl, Fadi Biadsy, and Shankar Kumar.
\newblock 2017.
\newblock Approaches for neural-network language model adaptation.
\newblock {\em Proc. Interspeech 2017}, pages 259--263.

\bibitem[\protect\citename{MacCartney}2005]{maccartney2005nlp}
Bill MacCartney.
\newblock 2005.
\newblock {NLP} lunch tutorial: Smoothing.

\bibitem[\protect\citename{Mass{\'o} and Badia}2010]{masso2010dealing}
Guillem Mass{\'o} and Toni Badia.
\newblock 2010.
\newblock Dealing with sign language morphemes in statistical machine
  translation.
\newblock In {\em 4th workshop on the representation and processing of sign
  languages: corpora and sign language technologies}.

\bibitem[\protect\citename{McTear}2004]{mctear:2004}
Mike McTear.
\newblock 2004.
\newblock {\em Spoken Dialogue Technology: Toward the Conversational User
  Interface}.
\newblock Springer, London.

\bibitem[\protect\citename{Merity \bgroup et al.\egroup }2017]{merityRegOpt}
Stephen Merity, Nitish~Shirish Keskar, and Richard Socher.
\newblock 2017.
\newblock {Regularizing and Optimizing LSTM Language Models}.
\newblock {\em arXiv preprint arXiv:1708.02182}.

\bibitem[\protect\citename{Merity \bgroup et al.\egroup }2018]{merityAnalysis}
Stephen Merity, Nitish~Shirish Keskar, and Richard Socher.
\newblock 2018.
\newblock {An Analysis of Neural Language Modeling at Multiple Scales}.
\newblock {\em arXiv preprint arXiv:1803.08240}.

\bibitem[\protect\citename{Mikolov \bgroup et al.\egroup }2009]{4960686}
T.~Mikolov, J.~Kopecky, L.~Burget, O.~Glembek, and J.~Cernocky.
\newblock 2009.
\newblock Neural network based language models for highly inflective languages.
\newblock In {\em 2009 IEEE International Conference on Acoustics, Speech and
  Signal Processing}, pages 4725--4728, April.

\bibitem[\protect\citename{Mikolov \bgroup et al.\egroup
  }2010]{mikolov2010recurrent}
Tom{\'a}{\v{s}} Mikolov, Martin Karafi{\'a}t, Luk{\'a}{\v{s}} Burget, Jan
  {\v{C}}ernock{\`y}, and Sanjeev Khudanpur.
\newblock 2010.
\newblock Recurrent neural network based language model.
\newblock In {\em Eleventh Annual Conference of the International Speech
  Communication Association}.

\bibitem[\protect\citename{Mocialov \bgroup et al.\egroup
  }2016]{mocialov2016towards}
Boris Mocialov, Patricia~A Vargas, and Micael~S Couceiro.
\newblock 2016.
\newblock Towards the evolution of indirect communication for social robots.
\newblock In {\em Computational Intelligence (SSCI), 2016 IEEE Symposium Series
  on}, pages 1--8. IEEE.

\bibitem[\protect\citename{Mocialov \bgroup et al.\egroup
  }2017]{mocialovtowards}
Boris Mocialov, Graham Turner, Katrin Lohan, and Helen Hastie.
\newblock 2017.
\newblock Towards continuous sign language recognition with deep learning.
\newblock In {\em Proc. of the Workshop on the Creating Meaning With Robot
  Assistants: The Gap Left by Smart Devices}.

\bibitem[\protect\citename{Muskan~Dhiman}2017]{muskandhimandrg.n.rathna2017}
Dr~G.N.~Rathna Muskan~Dhiman.
\newblock 2017.
\newblock Sign language recognition.
\newblock https://edu.authorcafe.com/academies/6813/sign-language-recognition.

\bibitem[\protect\citename{Rosenfeld}2000]{rosenfeld2000two}
Ronald Rosenfeld.
\newblock 2000.
\newblock Two decades of statistical language modeling: Where do we go from
  here?
\newblock {\em Proceedings of the IEEE}, 88(8):1270--1278.

\bibitem[\protect\citename{San-Segundo \bgroup et al.\egroup
  }2009]{san2009spoken}
Rub{\'e}n San-Segundo, Jos{\'e}~Manuel Pardo, Javier Ferreiros, Valent{\'\i}n
  Sama, Roberto Barra-Chicote, Juan~Manuel Lucas, D~S{\'a}nchez, and Antonio
  Garc{\'\i}a.
\newblock 2009.
\newblock Spoken spanish generation from sign language.
\newblock {\em Interacting with Computers}, 22(2):123--139.

\bibitem[\protect\citename{Schembri \bgroup et al.\egroup
  }2013]{schembri2013building}
Adam Schembri, Jordan Fenlon, Ramas Rentelis, Sally Reynolds, and Kearsy
  Cormier.
\newblock 2013.
\newblock Building the {British Sign Language} corpus.
\newblock {\em Language Documentation and Conservation 7}.

\bibitem[\protect\citename{Serban \bgroup et al.\egroup
  }2016]{serban2016generative}
Iulian~Vlad Serban, Ryan Lowe, Laurent Charlin, and Joelle Pineau.
\newblock 2016.
\newblock Generative deep neural networks for dialogue: A short review.
\newblock {\em arXiv preprint arXiv:1611.06216}.

\bibitem[\protect\citename{Stein \bgroup et al.\egroup }2007]{stein2007hand}
Daniel Stein, Philippe Dreuw, Hermann Ney, Sara Morrissey, and Andy Way.
\newblock 2007.
\newblock Hand in hand: automatic sign language to english translation.

\bibitem[\protect\citename{Stolcke}2002]{stolcke2002srilm}
Andreas Stolcke.
\newblock 2002.
\newblock Srilm-an extensible language modeling toolkit.
\newblock In {\em Proceedings of the Seventh international conference on spoken
  language processing}.

\bibitem[\protect\citename{Sutton-Spence and Woll}1999]{sutton1999linguistics}
R.~Sutton-Spence and B.~Woll.
\newblock 1999.
\newblock {\em The Linguistics of {British Sign Language}: An Introduction}.
\newblock The Linguistics of British Sign Language: An Introduction. Cambridge
  University Press.

\bibitem[\protect\citename{Yosinski \bgroup et al.\egroup
  }2014]{yosinski2014transferable}
Jason Yosinski, Jeff Clune, Yoshua Bengio, and Hod Lipson.
\newblock 2014.
\newblock How transferable are features in deep neural networks?
\newblock In {\em Advances in neural information processing systems}, pages
  3320--3328.

\bibitem[\protect\citename{Zhao \bgroup et al.\egroup }2000]{zhao2000machine}
Liwei Zhao, Karin Kipper, William Schuler, Christian Vogler, Norman Badler, and
  Martha Palmer.
\newblock 2000.
\newblock A machine translation system from english to american sign language.
\newblock In {\em Conference of the Association for Machine Translation in the
  Americas}, pages 54--67. Springer.

\end{thebibliography}

\end{document}